\documentclass[10pt,twocolumn,letterpaper]{article}
\usepackage[final]{cvpr}

\usepackage{amsmath,amsfonts,bm}

\def\eqref#1{equation~\ref{#1}}
\def\1{\bm{1}}

\DeclareMathAlphabet{\mathsfit}{\encodingdefault}{\sfdefault}{m}{sl}
\SetMathAlphabet{\mathsfit}{bold}{\encodingdefault}{\sfdefault}{bx}{n}

\usepackage{amsmath,amssymb}
\usepackage{booktabs}
\usepackage{graphicx}
\usepackage{microtype}
\definecolor{cvprblue}{rgb}{0.21,0.49,0.74}
\usepackage[pagebackref,breaklinks,colorlinks,allcolors=cvprblue]{hyperref}
\hypersetup{pdfauthor={},pdftitle={Counterfactual Velocity Steering for World Action Models}}

\def\confName{CVPR}
\def\confYear{2027}

\title{Counterfactual Velocity Steering for World Action Models}

\author{Xufeng Zhan, Yuanzhi Liang, Ye Lu, Haibin Huang, Chi Zhang, Xuelong Li}

\newcommand{\method}{GCAD}
\newcommand{\vtask}{v_{\mathrm{task}}}
\newcommand{\vmean}{v_{\mathrm{mean}}}
\newcommand{\vtail}{v_{\mathrm{tail}}}
\newcommand{\vcenterfield}{v_{\mathrm{center}}}
\newcommand{\vnoop}{v_{\mathrm{noop}}}

\newcommand{\orth}[2]{\operatorname{Orth}(#1;#2)}

\begin{document}

\maketitle

\begin{abstract}
Flow-based robot policies generate actions by repeatedly evaluating a conditional
velocity field. This makes test-time bias correction appear deceptively simple:
query the action expert under a task-agnostic or nominally stationary condition,
then subtract the resulting velocity from the task-conditioned prediction. We
show that this interpretation is often wrong. On Fast-WAM evaluated over 50
RoboTwin 2.0 tasks, predictions obtained from a mean instruction are highly
aligned with task-conditioned velocities (cosine approximately $0.94$--$0.95$).
Averaging eight completed-state references barely cancels this direction: the
norm of their mean is $98.3\%$ of the mean individual norm. Thus, reference
branches contain a large shared transport component, not only harmful bias, and
aggressive subtraction damages the generative trajectory. We develop a
same-state diagnostic framework and a shrinkage analysis showing that the
optimal correction scale depends on reference signal-to-noise. This motivates
\emph{geometry-calibrated action debiasing} (\method{}), which removes only the
component orthogonal to the task velocity and predicts a small, task-dependent
scale from within-task and cross-task instruction geometry. Controlled
RoboTwin experiments reveal three findings: corrupting the instruction still
permits $25.1\%$ success on a complete 50-task, 20-episode study, whereas
dynamic corruption of visual observations nearly eliminates success; a fixed centered
orthogonal correction reaches $92.7\%$ versus a $92.1\%$ baseline in one
randomized 1,000-trial sweep, but gains are non-monotonic and within observed
run-to-run variation; and an angular scale pilot has an offline oracle-gated
upper bound of $94.5\%$, while its best learned online realization reaches
$92.5\%$. Our results identify shared flow and train--serve mismatch as central
obstacles, and provide a reproducible foundation for conditional correction
rather than claiming that reference subtraction is universally beneficial.
\end{abstract}

\section{Introduction}

World Action Models (WAMs) have emerged as a promising framework for embodied intelligence by jointly modeling visual dynamics and robot actions. Unlike conventional vision-language-action policies that directly map observations and instructions to actions, WAMs also learn how the observed world changes during interaction. This additional objective can encourage representations of object motion, contact dynamics, spatial relations, and task progression. It may also broaden the sources of supervision available for robot learning. Videos collected from humans, other robots, or passive observations can reveal how objects move and how manipulation unfolds, even when action labels for the target robot are unavailable. Recent WAMs have demonstrated promising transfer across tasks, environments, and robot embodiments, while large-scale video representation models have shown that visual pretraining can support physical reasoning and downstream robot planning~\citep{ye2026worldactionmodels,assran2025vjepa2}. In parallel, the broader AI Flow perspective emphasizes that capable generative models must be coordinated across heterogeneous models, devices, and interaction contexts rather than treated as isolated predictors~\citep{an2025aiflow,shao2025aiflowedge}. These results suggest that video modeling can provide useful knowledge beyond what is available from narrowly collected robot demonstrations, making it an increasingly important component of general-purpose robot learning.

However, learning to predict visual futures and learning to select an appropriate action are not equivalent objectives. A video model must account for many changes in the visual input, including object appearance, texture, lighting, camera motion, and background structure. A control policy, in contrast, should preserve the information that determines the correct action under the current task and environment while ignoring changes that are irrelevant to control. Consequently, visually accurate prediction does not necessarily imply reliable action selection, causal physical understanding, or robust adaptation to changes in the robot and its surroundings. Recent evaluations show that video generation models can produce plausible visual futures while failing to extrapolate physical rules beyond familiar cases~\citep{kang2025physical,motamed2025physicsiq}. Fast-WAM further demonstrates that much of the benefit of video modeling may come from shaping the learned representation during training, since competitive action performance can be retained after explicit future-video generation is removed at inference~\citep{yuan2026fastwam}. These findings leave open a broader question about conditional control: after video and action learning are combined, does the policy adapt its action to the full conditioning state—including task intent, visual observations, robot proprioception, and interaction context—or does it primarily reproduce a behavior associated with the current observation? This question is closely related to a broader concern in generative-model post-training: optimizing a convenient surrogate, whether a likelihood term or a learned reward, can improve the measured objective while failing to preserve intended semantics, controllability, or robustness~\citep{liang2025rlvisual,liang2026teleboost}.

A potential failure mode is that WAM predictions become dominated by the current visual observation. When similar visual scenes repeatedly co-occur with particular actions in the training data, the policy may learn to reproduce the visually associated behavior while paying insufficient attention to the task instruction and robot state. Recent VLA studies report related modality imbalance, in which strong visual-action priors weaken the influence of language and proprioception~\citep{fei2025liberoplus,xu2025bayesvla}. This is also analogous to the non-visual statistical shortcuts exposed in visual-relationship recognition, where models can exploit predictable relationships without using the image evidence intended to support the decision~\citep{liang2019vrr}. We observe a similar asymmetry in Fast-WAM: replacing the instruction with one sampled from another task still yields $25.1\%$ success on randomized RoboTwin episodes, whereas dynamically replacing the visual observation reduces success to $1.5\%$ on the same condition. These results are consistent with the possibility that Fast-WAM relies strongly on visual observations and may not always respond adequately to non-visual conditioning.

In this paper, we propose a counterfactual velocity-steering framework for
diagnosing and mitigating conditioning imbalance in WAMs. Our starting point
is to query the action expert under alternative instruction contexts while
holding the visual observation, proprioception, noisy action, and flow time
fixed. A mean context constructed from training instructions serves as a
task-agnostic reference, allowing us to examine which components of the action
prediction persist when task-specific language is attenuated. This reference
is a probe, not a ground-truth estimate of bias: neither averaging embeddings
nor removing task intent guarantees that the resulting prediction should
correspond to a stationary robot.

The main challenge is to distinguish an unwanted action tendency from the
transport required by the generative process itself. Even a stationary action
endpoint generally requires a nonzero velocity at an intermediate noisy state.
Consistent with this distinction, our diagnostics reveal strong alignment
between task-conditioned velocities and both mean-instruction and
completed-state reference velocities. Directly subtracting a reference can
therefore remove useful transport together with any unwanted component. We
address this ambiguity by constructing centered residuals and retaining their
components orthogonal to the task velocity. This preserves the instantaneous
projection along the original prediction, while allowing a controlled change
in direction; it does not assume that the original prediction is always
correct or that the residual is purely harmful.

We instantiate this framework as \emph{geometry-calibrated action debiasing}
(\method{}). A noisy-reference analysis motivates shrinking the correction
according to its signal-to-noise ratio rather than applying a uniform, large
scale. We use within-task and cross-task instruction geometry as observable
proxies for this calibration, separating the correction amplitude from a gate
that estimates whether an intervention is beneficial. The same residual also provides a training-time signal: we use it to define
two detached sample-weighting rules based on angular displacement and relative
correction norm. These objectives test whether reference information can be
incorporated during learning without requiring an auxiliary branch at
deployment.
% Training-time gains remain unclaimed pending matched one-epoch evaluations.

Our RoboTwin study combines input interventions, same-state velocity
diagnostics, and controlled correction sweeps. The results show that reference
geometry is informative but does not make debiasing automatic: small
corrections help some tasks, while stronger corrections can hurt, and offline
scale-selection gains need not transfer to online rollouts. We therefore
distinguish evidence of conditional dependence, the quality of a candidate
bias estimate, and the downstream utility of applying it. Our contributions
are:
\begin{itemize}
    \item A controlled diagnostic framework that probes conditioning
    dependence in Fast-WAM and distinguishes action-space stationarity from
    shared transport in velocity space.
    \item A geometry-aware correction formulation combining centered
    residuals, orthogonal projection, and calibrated scales, together with
    training-time extensions based on corrected supervision and detached
    loss weighting.
    \item An empirical analysis of reference selection and correction
    strength on RoboTwin, identifying task-dependent benefits, failure modes
    of naive subtraction, and the gap between offline calibration and
    deployed control.
\end{itemize}

\section{Related Work}

\paragraph{Generative robot policies.}
Imitation-learning policies have progressed from direct action regression to
expressive sequence generators. Action Chunking with Transformers (ACT)
models temporally coherent action chunks for precise bimanual manipulation
\citep{zhao2023act}, while Diffusion Policy represents multimodal visuomotor
behavior through iterative action generation \citep{chi2023diffusionpolicy}.
Recent extensions incorporate explicit spatial structure, including 3D
Diffusion Policy and 3D Diffuser Actor, which combine action generation with
point-cloud or 3D scene representations and language or proprioceptive
conditioning \citep{ze2024dp3,ke2024diffuseractor}. These works motivate our
use of a conditional action field, but do not study how counterfactual
conditioning signals should be used to debias that field during fine-tuning.

\paragraph{Generalist vision-language-action policies.}
Large-scale robot datasets and multimodal pretraining have enabled policies
that transfer across tasks, embodiments, and environments. RT-1 studies
scaling a robotics transformer across diverse real-world data
\citep{brohan2022rt1}; RT-2 transfers web-scale vision-language knowledge into
robotic control by representing actions in the same sequence model
\citep{brohan2023rt2}. Open X-Embodiment and RT-X further investigate
cross-robot data and positive transfer \citep{openx2023}, while Octo provides
an open generalist policy intended for efficient downstream adaptation
\citep{octo2024}. OpenVLA uses an open vision-language backbone for
instruction-conditioned manipulation \citep{kim2024openvla}, and $\pi_0$
uses a flow-based action expert for general robot control \citep{black2024pi0}.
Our focus is complementary: given a WAM action expert that already receives
vision, language, and proprioception, we construct a same-state reference and
use its geometry to shape fine-tuning.

\paragraph{World models and controllable video generation.}
Classical world-model work learns predictive latent representations to support
planning and policy learning in imagined trajectories \citep{ha2018worldmodels}.
Video-pretrained representations such as R3M demonstrate that information
from human video can improve robot manipulation with limited robot data
\citep{nair2022r3m}; more recent self-supervised video world models extend this
line toward physical prediction and robot planning \citep{assran2025vjepa2}.
Fast-WAM couples video modeling with action generation while removing explicit
future-video imagination at inference \citep{yuan2026fastwam}. We retain this
world-action perspective but study a narrower failure mode: the action expert
may respond to a shared visual or transport direction even when the task
conditioning changes. RoboTwin 2.0 supplies a controlled benchmark for this
analysis through multi-task bimanual manipulation and strong scene
randomization \citep{chen2025robotwin}.

Controllable video generation provides a complementary perspective on the same
conditioning problem. VAST 1.0 separates textual scene planning from video
synthesis in order to improve controllability and temporal consistency
\citep{zhang2024vast}. In motion generation, SEEG explicitly separates
semantic and rhythm-related gesture factors and adds semantic-aware supervision
\citep{liang2022seeg}. These works show that a generative model can contain
both task-relevant and task-irrelevant motion components; our analysis asks how
the analogous components appear in a robot action velocity field.

\paragraph{Post-training and reward alignment for visual generation.}
Recent work has established reinforcement learning and preference optimization
as practical tools for post-training visual generators. A recent survey
organizes the foundations and design space of integrating RL with image, video,
and structured visual generation \citep{liang2025rlvisual}, while TeleBoost
emphasizes a staged post-training stack that combines supervised shaping,
reward-driven optimization, and preference refinement under stability
constraints \citep{liang2026teleboost}. Several methods focus on making the
resulting reward signal more reliable. Bayesian Prior-Guided Optimization
modulates trust in group updates using a semantic prior \citep{liu2026bpgo},
whereas TaRoS addresses reward saturation and target drift in video GRPO by
reorganizing multi-aspect reward components \citep{li2026taros}. ViPO further
replaces a single scalar visual advantage with structured spatial and temporal
feedback \citep{ni2026vipo}. These methods primarily optimize the quality or
preference of generated visual content; our method instead uses
counterfactual, same-state predictions to regularize a robot action field and
to diagnose conditioning imbalance.

\paragraph{Trajectory-level credit and adaptive guidance.}
The iterative nature of diffusion and flow-based generation has also motivated
step-aware alignment. Objective-aware Process Optimization assigns credit to
different denoising steps and adapts the combination of heterogeneous rewards
over the trajectory \citep{li2026otca}. Reward-Aware Trajectory Shaping uses
reward-dependent gates to decide when a few-step student should follow a
teacher trajectory \citep{li2026rats}. Our correction operates at the same
iterative field level, but has a different purpose: rather than transferring a
teacher trajectory or allocating visual rewards, it constructs a reference
velocity from counterfactual conditioning and applies a geometry-controlled
correction to the action prediction.

\paragraph{Bias diagnostics and adaptive supervision.}
VrR-VG demonstrates that a model can exploit relationships predictable from
non-visual statistics, motivating evaluation protocols that test whether a
prediction uses the information it is intended to use \citep{liang2019vrr}.
Similarly, episodic linear probing uses detached representations and adaptive
sample reweighting to measure and improve feature generalization during
training \citep{liang2022elp}. These ideas are conceptually related to our use
of counterfactual probes and detached correction weights. However, we study
continuous action-flow vectors in a multi-modal robot policy, and explicitly
separate shared transport from a potentially harmful conditioning residual.

\paragraph{Diffusion, flow matching, and conditional guidance.}
Denoising diffusion models learn to reverse a gradual corruption process
\citep{ho2020ddpm}, and score-based SDEs provide a continuous-time view of
noise-to-data generation \citep{song2021score}. Flow matching instead regresses
vector fields along prescribed probability paths \citep{lipman2023flow}, while
rectified flow emphasizes straighter transport paths that can be integrated
with fewer steps \citep{liu2022rectified}. Classifier-free guidance combines
conditional and unconditional predictions at sampling time
\citep{ho2022classifierfree}; related conditioning strategies have also been
used in diffusion-based robot policies. Our method differs in both objective
and timing: we do not claim an unconditional prediction or alter the sampler
at deployment. We use a counterfactual reference only to construct detached,
correction-aware weights during fine-tuning.

\section{Preliminaries and Diagnostic Setup}
\label{sec:preliminaries}

\subsection{Conditional action flow}

Let $o$ denote the current multi-view observation, $p$ the robot
proprioception, $c$ an instruction context, and
$a\in\mathbb{R}^{H\times D}$ a normalized action chunk. Fast-WAM learns a
conditional velocity field using flow matching. With Gaussian noise
$\epsilon$, scheduler time $t$, and its corresponding noise level
$\sigma_t\in[0,1]$, the training pair is
\begin{equation}
    \begin{aligned}
        x_t&=(1-\sigma_t)a+\sigma_t\epsilon,\qquad y=\epsilon-a,\\
        \vtask&=f_\theta(x_t,t;o,p,c).
    \end{aligned}
    \label{eq:action_flow}
\end{equation}
The network predicts velocity with respect to $\sigma$, not a physical joint
velocity. At inference, $\sigma$ decreases from one to zero and the scheduler
updates $x_{k+1}=x_k+(\sigma_{k+1}-\sigma_k)v_k$. We use the existing
noise schedule and action parameterization throughout.

For a binary mask $m$ over action tokens and dimensions, define the
masked mean squared error
\begin{equation}
    \mathcal{E}_{m}(u,y)
    =\frac{\sum_{h,d}m_{hd}(u_{hd}-y_{hd})^2}
           {\max(1,\sum_{h,d}m_{hd})}.
    \label{eq:masked_error}
\end{equation}
Writing $P_i$ for the original action-loss padding mask and $\omega_i$ for
the scheduler's timestep weight, the standard minibatch objective is
\begin{equation}
    \begin{aligned}
        \ell_i&=\omega_i\mathcal{E}_{P_i}(\vtask^{(i)},y_i),\\
        \mathcal{L}_{\mathrm{base}}
        &=\mathcal{L}_{\mathrm{video}}+\frac{1}{N}\sum_{i=1}^{N}\ell_i.
    \end{aligned}
    \label{eq:base_training}
\end{equation}
Here $\mathcal{L}_{\mathrm{video}}$ includes the original video-loss
weighting. The action and video objective coefficients are both one in our
fine-tuning experiments.

\subsection{Same-state reference predictions}

\paragraph{Mean-instruction reference.}
We construct a fixed mean context $\bar c$ from 8,192 cached training
instructions and query
\begin{equation}
    \vmean=f_\theta(x_t,t;o,p,\bar c).
    \label{eq:mean_reference}
\end{equation}
The task and reference branches share the same noisy action, flow time,
observation, and proprioception. Thus their difference is not confounded by
independently sampled action noise or separately integrated trajectories.
Averaging contexts weakens task-specific language without introducing an
empty string, but it is not guaranteed to produce an in-distribution null
instruction. We use this branch as a reference for debiasing, not as a
ground-truth bias label.

\paragraph{No-operation reference.}
Let $a_{\mathrm{noop}}(p)$ be a stationary action chunk in the policy's
action coordinates. For our absolute joint-position representation, we repeat
the current joint state over the horizon and apply the \emph{action}
normalizer; this need not equal the normalized proprioceptive input.
For $\sigma_t>0$, the corresponding analytic field is
\begin{equation}
    \vnoop=\frac{x_t-a_{\mathrm{noop}}(p)}{\sigma_t}.
    \label{eq:noop_reference}
\end{equation}
Although its endpoint is stationary, this velocity is generally nonzero.
Indeed, for the local clean-action estimate
$\hat a_{\mathrm{mean}}=x_t-\sigma_t\vmean$,
$\vmean-\vnoop=(a_{\mathrm{noop}}-\hat a_{\mathrm{mean}})/\sigma_t$.
This distinction motivates subtracting a reference \emph{residual} rather
than interpreting the entire reference velocity as unwanted motion.

\paragraph{Completed-state empirical center.}
Earlier diagnostic analyses also use a bank of matched terminal
frame--instruction pairs. Sampling $Q$ such pairs gives
\begin{equation}
    \vcenterfield=\vtail
    =\frac{1}{Q}\sum_{q=1}^{Q}
    f_\theta(x_t,t;o_q^{\mathrm{tail}},p,c_q^{\mathrm{tail}}).
    \label{eq:empirical_center}
\end{equation}
Each sampled frame is paired with its own instruction, and all branches
share $(x_t,t)$. The current proprioception is retained, so these visual and
proprioceptive inputs need not describe a consistent robot configuration.
We therefore distinguish this empirical center from the analytic no-op
field and evaluate it as an alternative reference, not an exact stationary
condition.

\section{Correction-Based Action Debiasing}
\label{sec:method}

We formulate action debiasing as a training-time signal for fine-tuning the
conditional action expert. The signal is constructed from same-state
counterfactual predictions and is used only to reweight the supervised
video--action objective. We study two loss designs: \emph{similarity-weighted
correction} and \emph{relative-norm-weighted correction}. Both use the same
detached correction geometry but assign training emphasis through different
measures of correction magnitude. At deployment, both fine-tuned models use
the unmodified Fast-WAM sampler; the experiments therefore isolate what can
be learned from correction-aware supervision rather than from additional
inference-time computation.

% Replace the previous method-overview figure block with:
% \input{figures/method_overview_v5.tex}
\begin{figure*}[t]
\centering
\includegraphics[width=\linewidth]{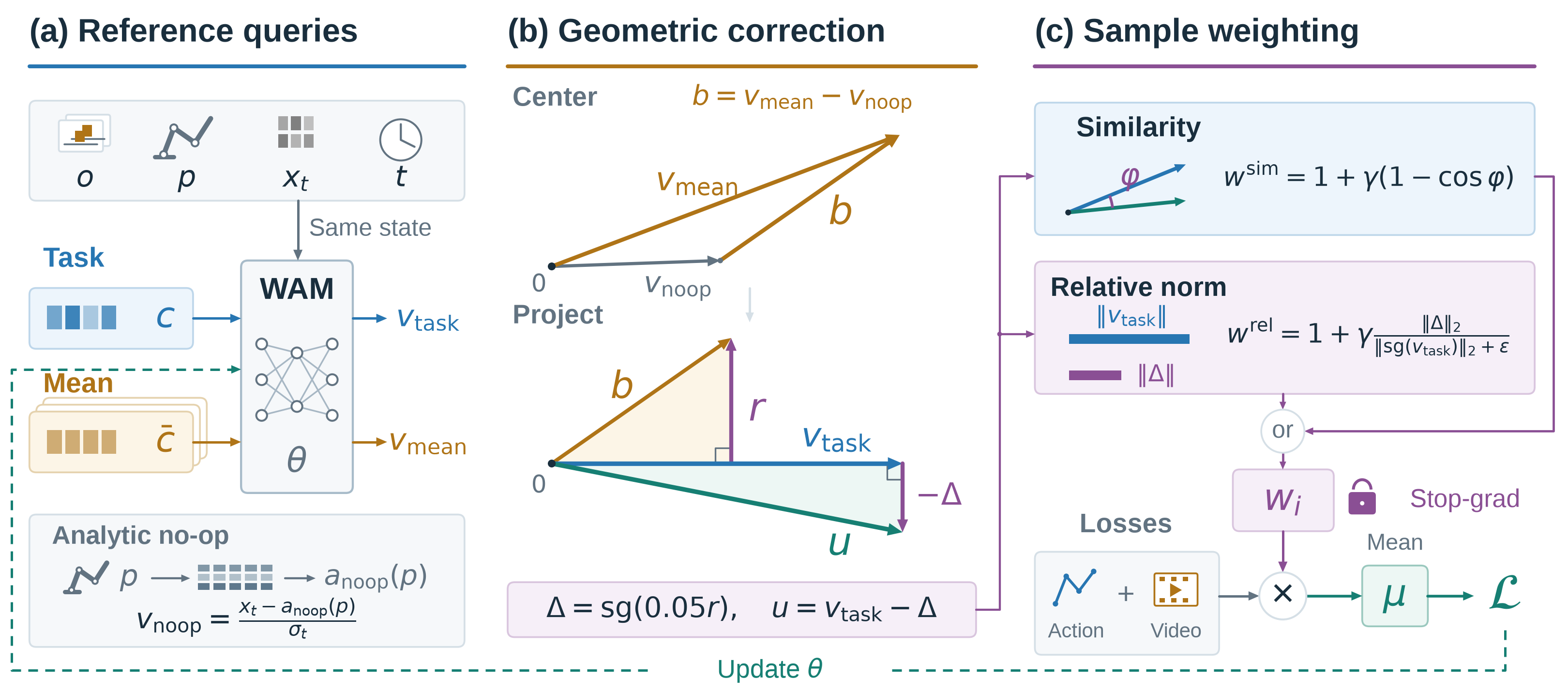}
\caption{\textbf{Correction-aware fine-tuning for Fast-WAM.}
(a) Task and mean instruction contexts query the same WAM with identical
observation $o$, proprioception $p$, noisy action $x_t$, and flow time $t$.
The mean context averages 8,192 cached training contexts. The no-op inset
repeats the current unnormalized joint positions over the action horizon and
applies the action normalizer to obtain $a_{\mathrm{noop}}(p)$. The analytic
no-op field is defined for $\sigma_t>0$
(Equation~\ref{eq:noop_reference}); a stationary endpoint generally requires
nonzero transport from the noisy state.
(b) The tip-to-tip arrow forms the centered residual
$b=v_{\mathrm{mean}}-v_{\mathrm{noop}}$. After masking padded and gripper
entries, orthogonal projection retains $r$. The detached correction
$\Delta=\operatorname{sg}(0.05r)$ yields $u=v_{\mathrm{task}}-\Delta$.
The geometry is shown in the valid non-gripper subspace, with the deflection
enlarged for visibility.
(c) The angle $\varphi$ between $v_{\mathrm{task}}$ and $u$, and the relative
correction norm, define two alternative sample-weighting rules (Equations~\ref{eq:similarity_weight}
and~\ref{eq:relative_weight}). The angle uses valid action entries; the
relative norm excludes gripper entries. Samples with $\sigma_t<0.25$ retain
unit weight. The lock denotes stop-gradient. The selected weight multiplies
the sum of the original action and video losses, and $\mu$ denotes the
minibatch mean. The dashed green path updates the shared model parameters.
The reference and weighting paths are used only during training; deployment
uses the original Fast-WAM sampler.}
\label{fig:method_overview}
\end{figure*}

\subsection{Same-state correction construction}
\label{sec:correction_operator}

For each training sample, we evaluate a task branch and a mean-reference
branch at the same noisy action state, flow time, observation, and
proprioception. The reference instruction is the mean of 8,192 cached
training instructions. We use the analytic stationary-endpoint field from
Equation~\ref{eq:noop_reference} to remove the part of the reference that is
explained by the current no-operation endpoint. Thus, for sample $i$,
\begin{equation}
    b_i=\vmean^{(i)}-\vnoop^{(i)}.
    \label{eq:training_residual}
\end{equation}
This construction is state matched: the residual is not obtained from a
separately integrated trajectory or an unrelated observation. The no-operation
field is generally nonzero in velocity space because a stationary endpoint
still requires transport from the current noisy state.

The residual may nevertheless share a component with the useful generative
transport direction. We therefore retain only the component orthogonal to the
task prediction. Let $M_i$ be a mask over valid action tokens and dimensions,
$\tilde b_i=M_i\odot b_i$, and
$\tilde v_i=M_i\odot\vtask^{(i)}$. We define
\begin{equation}
    \begin{aligned}
    r_i&=\orth{\tilde b_i}{\tilde v_i}\\
       &=\tilde b_i-
        \frac{\langle\tilde b_i,\tilde v_i\rangle}
             {\max(\|\tilde v_i\|_2^2,\delta)}\tilde v_i.
    \end{aligned}
    \label{eq:training_orthogonal}
\end{equation}
where $\delta=10^{-8}$ stabilizes the division, and padded entries and
gripper dimensions are excluded from the projection. For nondegenerate
$\tilde v_i$, this residual has zero inner product with the task direction.
The detached correction used by both objectives is
\begin{equation}
    \Delta_i=\operatorname{sg}(s r_i),
    \qquad
    u_i=\vtask^{(i)}-\Delta_i,
    \label{eq:detached_correction}
\end{equation}
with fixed $s=0.05$. The stop-gradient operator prevents the reference query
and the projection geometry from introducing an additional competing gradient
path; gradients from the weighted objective flow through the task prediction
and the shared model parameters only.

\subsection{Correction-aware loss weighting}
\label{sec:correction_weighting}

For each sample, let $A_i$ and $V_i$ denote the original masked action and
video losses, respectively, and let $\lambda_A$ and $\lambda_V$ be their
fixed coefficients. The standard objective is
\begin{equation}
    \mathcal{L}_{\mathrm{base}}
    =\frac{1}{N}\sum_{i=1}^{N}
      \left(\lambda_A A_i+\lambda_V V_i\right).
    \label{eq:base_objective_finetune}
\end{equation}
We apply correction-derived weighting only to samples with
$\sigma_{t_i}\geq\sigma_{\min}=0.25$; all other samples retain unit weight.
The weights are detached, so the correction changes the emphasis assigned to
a sample without creating a second gradient path through the reference
branch.

\paragraph{Similarity-weighted correction.}
This objective measures how much the corrected prediction changes the task
velocity in direction. Define
\begin{equation}
    d_i=1-\cos\!\left(\vtask^{(i)},u_i\right),
    \qquad
    w_i^{\mathrm{sim}}=1+\gamma d_i,
    \label{eq:similarity_weight}
\end{equation}
for eligible samples and set $w_i^{\mathrm{sim}}=1$ otherwise. The objective is
\begin{equation}
    \mathcal{L}_{\mathrm{sim}}=\frac{1}{N}\sum_{i=1}^{N}w_i^{\mathrm{sim}}
      \left(\lambda_A A_i+\lambda_V V_i\right).
    \label{eq:similarity_objective}
\end{equation}
The cosine is computed over valid action entries after masking padded
positions. This weighting emphasizes samples for which the correction changes
the task prediction direction more substantially.

\paragraph{Relative-norm-weighted correction.}
This objective measures the correction magnitude relative to the task
prediction. With $\Delta_i$ from Equation~\ref{eq:detached_correction}, define
\begin{equation}
    \rho_i=\frac{\|\Delta_i\|_2}
                 {\|\operatorname{sg}(\vtask^{(i)})\|_2+\epsilon},
    \qquad
    w_i^{\mathrm{rel}}=1+\gamma\rho_i,
    \label{eq:relative_weight}
\end{equation}
for eligible samples and $w_i^{\mathrm{rel}}=1$ otherwise. The objective is
\begin{equation}
    \mathcal{L}_{\mathrm{rel}}=\frac{1}{N}\sum_{i=1}^{N}w_i^{\mathrm{rel}}
      \left(\lambda_A A_i+\lambda_V V_i\right).
    \label{eq:relative_objective}
\end{equation}
The denominator is evaluated over valid non-gripper action dimensions. The
two objectives share the same correction construction and differ only in the
statistic that determines the detached sample weight.

\paragraph{Training protocol.}
The two correction-aware objectives are initialized independently from the
released Fast-WAM checkpoint and trained for 5,000 optimizer steps with global
batch size 128, initial learning rate $10^{-5}$, cosine scheduling, and seed
42. We use $s=0.05$, $\sigma_{\min}=0.25$, and the same video--action
architecture, normalization, and data pipeline as standard fine-tuning. The
standard continued-training model is the matched control. At evaluation time,
all models use the original sampler with inference correction disabled.

\section{Experiments}
\label{sec:experiments}

\subsection{Conditioning dependence under input interventions}
\label{sec:conditioning_interventions}

We first examine whether Fast-WAM responds to each conditioning modality or
can complete a task from a strong visual-action prior alone. We intervene on
one input at a time while keeping the remaining inputs unchanged. In a
wrong-instruction episode, the task instruction is replaced by an instruction
sampled from a different task, while the observation and robot state remain
correct. In a dynamic wrong-observation episode, the correct instruction is
retained, but the observation is replaced by synchronized frames from an
unrelated training episode. The donor frames are replayed progressively over
the rollout rather than freezing one image; proprioception remains the live
value from the evaluated episode. We also report a stale-observation control
that repeats the correct initial frame throughout the episode.

The interventions are evaluated on both RoboTwin and LIBERO. RoboTwin reports
clean and randomized conditions separately; the updated intervention study
uses 20 episodes per task and completes all 50 tasks. LIBERO uses four suites
and 800 episodes with a fixed seed. All intervention results use the original
released checkpoint and disable inference-time correction.

\begin{table*}[t]
\centering
\caption{Input intervention studies. Rows denote the three input conditions:
the correct-input baseline, a wrong instruction, and a dynamic wrong
observation. Columns denote evaluation categories: clean and random for
RoboTwin, and long, spatial, object, and goal for LIBERO. RoboTwin intervention
columns use 20 episodes per task and cover all 50 tasks, while the baseline is
the matched full evaluation. LIBERO reports suite-level success rates over 200
episodes per suite. All values are percentages.}
\label{tab:conditioning_interventions}
\small
\begin{tabular*}{\textwidth}{@{\extracolsep{\fill}}lrrrrrr@{}}
\toprule
& \multicolumn{2}{c}{RoboTwin} & \multicolumn{4}{c}{LIBERO} \\
\cmidrule(lr){2-3} \cmidrule(lr){4-7}
Condition & Clean & Random & Long & Spatial & Object & Goal \\
\midrule
Baseline & 92.16 & 91.94 & 94.00 & 97.50 & 98.50 & 96.50 \\
Wrong instruction & 31.30 & 25.10 & 30.50 & 24.00 & 64.00 & 3.00 \\
Dynamic wrong observation & 2.00 & 1.50 & 0.00 & 0.00 & 0.00 & 0.00 \\
\bottomrule
\end{tabular*}
\end{table*}

The two benchmarks exhibit the same qualitative asymmetry. Replacing the
instruction does not eliminate all successful behavior: the updated RoboTwin
study retains $25.1\%$ randomized success, while LIBERO retains an overall
$30.375\%$ success rate, with the largest residual success in the object
suite. In contrast, dynamic observation replacement reduces RoboTwin success
to $2.0\%$ on clean scenes and $1.5\%$ under randomization, and reduces LIBERO
success to zero in every suite. The previously measured fixed-initial-frame
control also produced near-zero RoboTwin performance, showing that the policy
depends on temporally updated visual evidence rather than only on the
first-frame object layout. These results are consistent with a strong
observation-dependent action prior and substantial instruction insensitivity,
but they should not be interpreted as proving that language is unused: the
wrong-instruction condition still changes success, and the intervention
protocols differ in sample size and benchmark structure.

\subsection{RoboTwin fine-tuning}
\label{sec:robotwin_finetuning}

We evaluate correction-aware fine-tuning on RoboTwin 2.0, which contains 50
bimanual manipulation tasks under clean and randomized scene conditions. For
every model, we use 100 episodes per task and condition, a fixed start seed of
zero, a 32-step action horizon, ten flow-integration steps, and replanning
every 24 environment steps. The same released Fast-WAM checkpoint initializes
all fine-tuning runs. Each run uses 5,000 optimizer steps, global batch size
128, learning rate $10^{-5}$ with cosine scheduling, and seed 42. Inference-
time correction is disabled for all rows, including the correction-trained
models.

\begin{table*}[t]
\centering
\caption{RoboTwin fine-tuning results. Success rates are averaged over 50
tasks and 100 episodes per task in each condition. The overall average gives
equal weight to clean and randomized evaluation.}
\label{tab:training_robotwin}
\small
\begin{tabular*}{\textwidth}{@{\extracolsep{\fill}}lrrrrr@{}}
\toprule
Training variant & $\gamma$ & Clean & Randomized & Average & $\Delta$ vs. control \\
\midrule
Standard fine-tuning control & -- & 92.36 & 91.50 & 91.93 & -- \\
Similarity-weighted correction & 5 & 93.20 & 91.94 & 92.57 & +0.64 \\
Similarity-weighted correction & 10 & 92.54 & 91.90 & 92.22 & +0.29 \\
Relative-norm-weighted correction & 5 & 93.54 & 91.60 & 92.57 & +0.64 \\
Relative-norm-weighted correction & 2.5 & 92.48 & 92.14 & 92.31 & +0.38 \\
\bottomrule
\end{tabular*}
\end{table*}

\begin{figure*}[t]
\centering
\includegraphics[width=\linewidth]{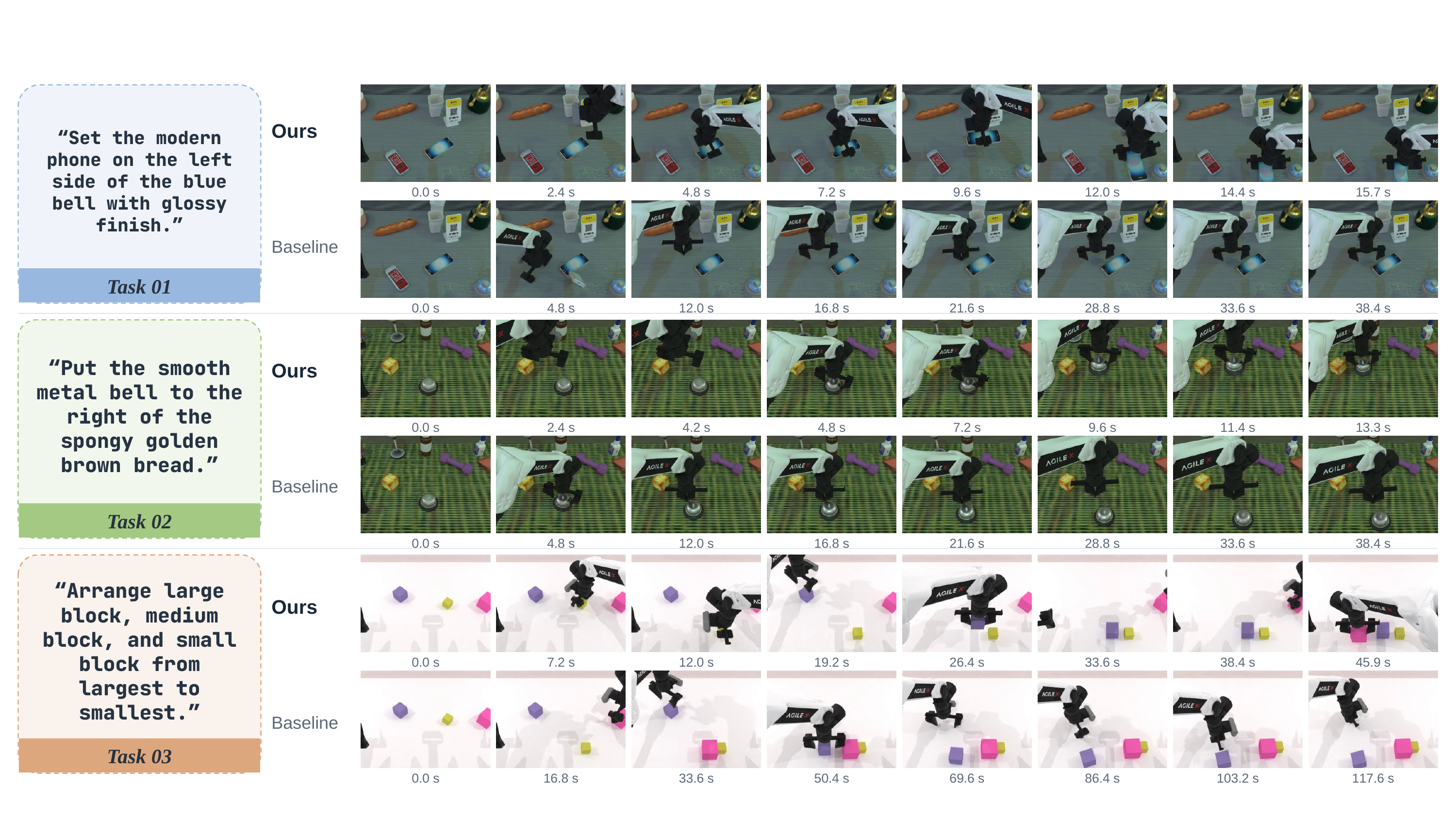}
\caption{Qualitative RoboTwin rollouts on three representative tasks.
Each row compares the correction-aware model (Ours) with the standard Fast-WAM
baseline under the same task instruction and initial scene. The frame
sequences illustrate how the two policies differ in object selection,
placement, and task completion.}
\label{fig:robotwin_visual}
\end{figure*}

Both correction-aware objectives improve the matched control by $0.64$\,pp
in the overall average at $\gamma=5$. The gain is driven primarily by clean
episodes: the similarity-weighted objective reaches $93.20\%$ and the
relative-norm-weighted objective reaches $93.54\%$, while their randomized
success rates remain close to the control. Increasing the similarity weight
from $\gamma=5$ to $10$ reduces the average improvement, and the relative-norm
objective shows the same non-monotonic behavior when its weight is reduced to
$2.5$. This pattern is consistent with the intended role of the correction
signal: it focuses learning on samples for which the reference geometry
indicates a substantial intervention, but excessive emphasis can distort the
original video--action objective.

\subsection{LIBERO-Plus language-instruction difficulty}
\label{sec:libero_plus_language}

We further evaluate whether correction-aware fine-tuning improves robustness
to difficult language instructions. We use the language-instruction split of
LIBERO-Plus and evaluate 1,537 instruction variants, one episode per variant,
with a fixed seed. The benchmark assigns each variant to one of five empirical
difficulty levels according to how many of four reference policies solve it:
Level 1 contains variants solved by all four policies, whereas Level 5 contains
variants solved by none. Thus, higher levels represent increasingly difficult
language-instruction conditions rather than different rewriting operators.
Table~\ref{tab:libero_plus_language} reports the pooled success rate within
each level.

\begin{table*}[t]
\centering
\caption{LIBERO-Plus language-instruction difficulty breakdown. Higher levels
contain harder instruction variants. Best results in each level are bold.}
\label{tab:libero_plus_language}
\small
\begin{tabular*}{\textwidth}{@{\extracolsep{\fill}}lrrrrrr@{}}
\toprule
Difficulty & Variants & Baseline & \shortstack{Similarity\\$\gamma=5$} & \shortstack{Similarity\\$\gamma=10$} & \shortstack{Relative norm\\$\gamma=2.5$} & \shortstack{Relative norm\\$\gamma=5$} \\
\midrule
Level 1 & 328 & 81.10 & 78.35 & 78.05 & 78.66 & 78.66 \\
Level 2 & 280 & 73.93 & 75.36 & 71.79 & 71.79 & 75.36 \\
Level 3 & 259 & 73.75 & 73.75 & 74.13 & 74.52 & \textbf{74.90} \\
Level 4 & 328 & 67.38 & 66.16 & 67.07 & 66.16 & 66.16 \\
Level 5 & 342 & 46.49 & 46.78 & \textbf{49.42} & 48.54 & 48.25 \\
\bottomrule
\end{tabular*}
\end{table*}

The correction-aware models are particularly competitive on the harder
language strata. On Level 3, the relative-norm-weighted objective with
$\gamma=5$ reaches $74.90\%$, improving over the baseline by $1.15$\,pp. On
the most difficult Level 5, the similarity-weighted objective with $\gamma=10$
reaches $49.42\%$, a $2.93$\,pp improvement over the baseline. These results
indicate that the correction signal can improve robustness when the language
instruction is difficult, even though the gain is not uniform across all
levels: the same models lose performance on Level 1, and the improvement on
the hard strata does not translate into a reliable aggregate gain. This
pattern is consistent with correction-aware fine-tuning acting as a selective
regularizer rather than a universally beneficial boost.

\section{Discussion and Limitations}

The RoboTwin results support correction-aware fine-tuning as a modest but
measurable intervention. Both correction-aware objectives improve the matched
continued-training control, while the non-monotonic dependence on $\gamma$
indicates that the correction signal should not be treated as an unconditional
reward. The improvement is larger on clean episodes than on randomized
episodes, suggesting that the learned weighting primarily changes behavior in
familiar visual configurations. This result is consistent with the intended
use of the reference branch: it identifies examples whose conditional action
prediction is sensitive to a shared nuisance direction, but it does not
provide a complete task-independent error label.

The LIBERO-Plus analysis provides complementary evidence that the correction
signal is useful under difficult language conditions. The strongest variants
improve Level 3 and Level 5 success by $1.15$ and $2.93$\,pp, respectively,
although the gains are offset by degradation on easy Level 1 instructions.
This supports a selective-robustness interpretation rather than a uniform
improvement across the full instruction distribution.

The main limitation is that the two objectives share the same reference
construction and fixed correction scale, so the experiments do not separate
improvements from the choice of reference, projection, and weighting statistic.
The reference branch also adds computation during training. A practical
implementation should cache or amortize these queries and should verify that
detached weights do not distort the balance between video and action
supervision.

\section{Conclusion}

We presented correction-aware fine-tuning for Fast-WAM with two loss designs.
Both construct a same-state, orthogonal, detached correction from a
mean-instruction reference and use it to weight the original video--action
objective. On the completed RoboTwin comparison, both objectives improve the
matched standard fine-tuning control by $0.64$ percentage points at their best
tested setting. The result supports correction as a useful training signal
while also showing that its strength must be calibrated. The LIBERO-Plus
breakdown further suggests that the correction is most useful for difficult
language instructions, where it improves the best Level 3 and Level 5 results
without producing a uniform gain on easy cases. These findings provide a
focused basis for studying more efficient reference estimation and selective
robustness.

\section*{LLM Usage Statement}
OpenAI Codex (GPT-5-based) was used interactively during research and
manuscript preparation. Assistance included discussing experimental designs,
developing and debugging training, evaluation, and analysis scripts, preparing
visualizations and figure-generation code, and drafting and revising prose and
LaTeX throughout the manuscript. Its role was assistive; the authors retain
responsibility for the experimental evidence, citations, scientific claims,
and final submission.

{
\small
\bibliographystyle{ieeenat_fullname}
\bibliography{references}

@inproceedings{lipman2023flow,
  title = {Flow Matching for Generative Modeling},
  author = {Lipman, Yaron and Chen, Ricky T. Q. and Ben-Hamu, Heli and Nickel, Maximilian and Le, Matt},
  booktitle = {International Conference on Learning Representations},
  year = {2023},
  url = {https://arxiv.org/abs/2210.02747}
}

@inproceedings{chi2023diffusionpolicy,
  title = {Diffusion Policy: Visuomotor Policy Learning via Action Diffusion},
  author = {Chi, Cheng and Feng, Siyuan and Du, Yilun and Xu, Zhenjia and Cousineau, Eric and Burchfiel, Benjamin and Song, Shuran},
  booktitle = {Proceedings of Robotics: Science and Systems},
  year = {2023},
  doi = {10.15607/RSS.2023.XIX.026},
  url = {https://www.roboticsproceedings.org/rss19/p026.html}
}

@article{ho2022classifierfree,
  title = {Classifier-Free Diffusion Guidance},
  author = {Ho, Jonathan and Salimans, Tim},
  journal = {arXiv preprint arXiv:2207.12598},
  year = {2022},
  url = {https://arxiv.org/abs/2207.12598}
}

@inproceedings{black2024pi0,
  title = {$\pi_0$: A Vision-Language-Action Flow Model for General Robot Control},
  author = {Black, Kevin and Brown, Noah and Driess, Danny and Esmail, Adnan and Equi, Michael and Finn, Chelsea and Fusi, Niccolo and Groom, Lachy and Hausman, Karol and Ichter, Brian and Jakubczak, Szymon and Jones, Tim and Ke, Liyiming and Levine, Sergey and Li-Bell, Adrian and Mothukuri, Mohith and Nair, Suraj and Pertsch, Karl and Shi, Lucy Xiaoyang and Tanner, James and Vuong, Quan and Walling, Anna and Wang, Haohuan and Zhilinsky, Ury},
  booktitle = {Proceedings of Robotics: Science and Systems},
  year = {2025},
  doi = {10.15607/RSS.2025.XXI.010},
  url = {https://www.roboticsproceedings.org/rss21/p010.html}
}

@inproceedings{kim2024openvla,
  title = {{OpenVLA}: An Open-Source Vision-Language-Action Model},
  author = {Kim, Moo Jin and Pertsch, Karl and Karamcheti, Siddharth and Xiao, Ted and Balakrishna, Ashwin and Nair, Suraj and Rafailov, Rafael and Foster, Ethan P. and Sanketi, Pannag R. and Vuong, Quan and Kollar, Thomas and Burchfiel, Benjamin and Tedrake, Russ and Sadigh, Dorsa and Levine, Sergey and Liang, Percy and Finn, Chelsea},
  booktitle = {Proceedings of The 8th Conference on Robot Learning},
  volume = {270},
  pages = {2679--2713},
  year = {2025},
  publisher = {PMLR},
  url = {https://proceedings.mlr.press/v270/kim25c.html}
}

@article{chen2025robotwin,
  title = {{RoboTwin 2.0}: A Scalable Data Generator and Benchmark with Strong Domain Randomization for Robust Bimanual Robotic Manipulation},
  author = {Chen, Tianxing and Chen, Zanxin and Chen, Baijun and Cai, Zijian and Liu, Yibin and Li, Zixuan and Liang, Qiwei and Lin, Xianliang and Ge, Yiheng and Gu, Zhenyu and Deng, Weiliang and Guo, Yubin and Nian, Tian and Xie, Xuanbing and Chen, Qiangyu and Su, Kailun and Xu, Tianling and Liu, Guodong and Hu, Mengkang and Gao, Huan-ang and Wang, Kaixuan and Liang, Zhixuan and Qin, Yusen and Yang, Xiaokang and Luo, Ping and Mu, Yao},
  journal = {arXiv preprint arXiv:2506.18088},
  year = {2025},
  url = {https://arxiv.org/abs/2506.18088}
}

@article{yuan2026fastwam,
  title = {{Fast-WAM}: Do World Action Models Need Test-time Future Imagination?},
  author = {Yuan, Tianyuan and Dong, Zibin and Liu, Yicheng and Zhao, Hang},
  journal = {arXiv preprint arXiv:2603.16666},
  year = {2026},
  url = {https://arxiv.org/abs/2603.16666}
}

@inproceedings{ho2020ddpm,
  title = {Denoising Diffusion Probabilistic Models},
  author = {Ho, Jonathan and Jain, Ajay and Abbeel, Pieter},
  booktitle = {Advances in Neural Information Processing Systems},
  volume = {33},
  pages = {6840--6851},
  year = {2020},
  url = {https://proceedings.neurips.cc/paper/2020/hash/4c5bcfec8584af0d967f1ab10179ca4b-Abstract.html}
}

@inproceedings{song2021score,
  title = {Score-Based Generative Modeling through Stochastic Differential Equations},
  author = {Song, Yang and Sohl-Dickstein, Jascha and Kingma, Diederik P. and Kumar, Abhishek and Ermon, Stefano and Poole, Ben},
  booktitle = {International Conference on Learning Representations},
  year = {2021},
  url = {https://arxiv.org/abs/2011.13456}
}

@inproceedings{liu2022rectified,
  title = {Flow Straight and Fast: Learning to Generate and Transfer Data with Rectified Flow},
  author = {Liu, Xingchao and Gong, Chengyue and Liu, Qiang},
  booktitle = {International Conference on Learning Representations},
  year = {2023},
  url = {https://arxiv.org/abs/2209.03003}
}

@inproceedings{zhao2023act,
  title = {Learning Fine-Grained Bimanual Manipulation with Low-Cost Hardware},
  author = {Zhao, Tony Z. and Kumar, Vikash and Levine, Sergey and Finn, Chelsea},
  booktitle = {Proceedings of Robotics: Science and Systems},
  year = {2023},
  doi = {10.15607/RSS.2023.XIX.016},
  url = {https://www.roboticsproceedings.org/rss19/p016.html}
}

@inproceedings{brohan2022rt1,
  title = {{RT-1}: Robotics Transformer for Real-World Control at Scale},
  author = {Brohan, Anthony and Brown, Noah and Carbajal, Justice and Chebotar, Yevgen and Dabis, Joseph and Finn, Chelsea and Gopalakrishnan, Keerthana and Hausman, Karol and Herzog, Alexander and Hsu, Jasmine and Ibarz, Julian and Ichter, Brian and Irpan, Alex and Jackson, Tomas and Jesmonth, Sally and Joshi, Nikhil J. and Julian, Ryan and Kalashnikov, Dmitry and Kuang, Yuheng and Leal, Isabel and Lee, Kuang-Huei and Levine, Sergey and Lu, Yao and Malla, Utsav and Manjunath, Deeksha and Mordatch, Igor and Nachum, Ofir and Parada, Carolina and Peralta, Jodilyn and Perez, Emily and Pertsch, Karl and Quiambao, Jornell and Rao, Kanishka and Ryoo, Michael S. and Salazar, Grecia and Sanketi, Pannag R. and Sayed, Kevin and Singh, Jaspiar and Sontakke, Sumedh and Stone, Austin and Tan, Clayton and Tran, Huong T. and Vanhoucke, Vincent and Vega, Steve and Vuong, Quan and Xia, Fei and Xiao, Ted and Xu, Peng and Xu, Sichun and Yu, Tianhe and Zitkovich, Brianna},
  booktitle = {Proceedings of Robotics: Science and Systems},
  year = {2023},
  doi = {10.15607/RSS.2023.XIX.025},
  url = {https://doi.org/10.15607/RSS.2023.XIX.025}
}

@inproceedings{brohan2023rt2,
  title = {{RT-2}: Vision-Language-Action Models Transfer Web Knowledge to Robotic Control},
  author = {Zitkovich, Brianna and Yu, Tianhe and Xu, Sichun and Xu, Peng and Xiao, Ted and Xia, Fei and Wu, Jialin and Wohlhart, Paul and Welker, Stefan and Wahid, Ayzaan and Vuong, Quan and Vanhoucke, Vincent and Tran, Huong and Soricut, Radu and Singh, Anikait and Singh, Jaspiar and Sermanet, Pierre and Sanketi, Pannag R. and Salazar, Grecia and Ryoo, Michael S. and Reymann, Krista and Rao, Kanishka and Pertsch, Karl and Mordatch, Igor and Michalewski, Henryk and Lu, Yao and Levine, Sergey and Lee, Lisa and Lee, Tsang-Wei Edward and Leal, Isabel and Kuang, Yuheng and Kalashnikov, Dmitry and Julian, Ryan and Joshi, Nikhil J. and Irpan, Alex and Ichter, Brian and Hsu, Jasmine and Herzog, Alexander and Hausman, Karol and Gopalakrishnan, Keerthana and Fu, Chuyuan and Florence, Pete and Finn, Chelsea and Dubey, Kumar Avinava and Driess, Danny and Ding, Tianli and Choromanski, Krzysztof Marcin and Chen, Xi and Chebotar, Yevgen and Carbajal, Justice and Brown, Noah and Brohan, Anthony and Arenas, Montserrat Gonzalez and Han, Kehang},
  booktitle = {Proceedings of The 7th Conference on Robot Learning},
  volume = {229},
  pages = {2165--2183},
  year = {2023},
  publisher = {PMLR},
  url = {https://proceedings.mlr.press/v229/zitkovich23a.html}
}

@inproceedings{openx2023,
  title = {{Open X-Embodiment}: Robotic Learning Datasets and {RT-X} Models},
  author = {{Open X-Embodiment Collaboration}},
  booktitle = {2024 IEEE International Conference on Robotics and Automation (ICRA)},
  pages = {6892--6903},
  year = {2024},
  doi = {10.1109/ICRA57147.2024.10611477},
  url = {https://doi.org/10.1109/ICRA57147.2024.10611477}
}

@inproceedings{octo2024,
  title = {Octo: An Open-Source Generalist Robot Policy},
  author = {Ghosh, Dibya and Walke, Homer Rich and Pertsch, Karl and Black, Kevin and Mees, Oier and Dasari, Sudeep and Hejna, Joey and Kreiman, Tobias and Xu, Charles and Luo, Jianlan and Tan, You Liang and Chen, Lawrence Yunliang and Vuong, Quan and Xiao, Ted and Sanketi, Pannag R. and Sadigh, Dorsa and Finn, Chelsea and Levine, Sergey},
  booktitle = {Proceedings of Robotics: Science and Systems},
  year = {2024},
  doi = {10.15607/RSS.2024.XX.090},
  url = {https://www.roboticsproceedings.org/rss20/p090.html}
}

@inproceedings{ke2024diffuseractor,
  title = {{3D Diffuser Actor}: Policy Diffusion with {3D} Scene Representations},
  author = {Ke, Tsung-Wei and Gkanatsios, Nikolaos and Fragkiadaki, Katerina},
  booktitle = {Proceedings of The 8th Conference on Robot Learning},
  volume = {270},
  pages = {1949--1974},
  year = {2025},
  publisher = {PMLR},
  url = {https://proceedings.mlr.press/v270/ke25a.html}
}

@inproceedings{ze2024dp3,
  title = {{3D Diffusion Policy}: Generalizable Visuomotor Policy Learning via Simple {3D} Representations},
  author = {Ze, Yanjie and Zhang, Gu and Zhang, Kangning and Hu, Chenyuan and Wang, Muhan and Xu, Huazhe},
  booktitle = {Proceedings of Robotics: Science and Systems},
  year = {2024},
  url = {https://www.roboticsproceedings.org/rss20/p067.html}
}

@inproceedings{nair2022r3m,
  title = {{R3M}: A Universal Visual Representation for Robot Manipulation},
  author = {Nair, Suraj and Rajeswaran, Aravind and Kumar, Vikash and Finn, Chelsea and Gupta, Abhinav},
  booktitle = {Proceedings of The 6th Conference on Robot Learning},
  volume = {205},
  pages = {892--909},
  year = {2023},
  publisher = {PMLR},
  url = {https://proceedings.mlr.press/v205/nair23a.html}
}

@article{ha2018worldmodels,
  title = {World Models},
  author = {Ha, David and Schmidhuber, Jürgen},
  journal = {arXiv preprint arXiv:1803.10122},
  year = {2018},
  url = {https://arxiv.org/abs/1803.10122}
}

@article{assran2025vjepa2,
  title = {{V-JEPA 2}: Self-Supervised Video Models Enable Understanding, Prediction and Planning},
  author = {Assran, Mido and Bardes, Adrien and Fan, David and Garrido, Quentin and Howes, Russell and Komeili, Mojtaba and Muckley, Matthew and Rizvi, Ammar and Roberts, Claire and Sinha, Koustuv and Zholus, Artem and Arnaud, Sergio and Gejji, Abha and Martin, Ada and Hogan, Francois Robert and Dugas, Daniel and Bojanowski, Piotr and Khalidov, Vasil and Labatut, Patrick and Massa, Francisco and Szafraniec, Marc and Krishnakumar, Kapil and Li, Yong and Ma, Xiaodong and Chandar, Sarath and Meier, Franziska and LeCun, Yann and Rabbat, Michael and Ballas, Nicolas},
  journal = {arXiv preprint arXiv:2506.09985},
  year = {2025},
  url = {https://arxiv.org/abs/2506.09985}
}

@article{ye2026worldactionmodels,
  title = {World Action Models are Zero-shot Policies},
  author = {Ye, Seonghyeon and Ge, Yunhao and Zheng, Kaiyuan and Gao, Shenyuan and Yu, Sihyun and Kurian, George and Indupuru, Suneel and Tan, You Liang and Zhu, Chuning and Xiang, Jiannan and Malik, Ayaan and Lee, Kyungmin and Liang, William and Ranawaka, Nadun and Gu, Jiasheng and Xu, Yinzhen and Wang, Guanzhi and Hu, Fengyuan and Narayan, Avnish and Bjorck, Johan and Wang, Jing and Kim, Gwanghyun and Niu, Dantong and Zheng, Ruijie and Xie, Yuqi and Wu, Jimmy and Wang, Qi and Julian, Ryan and Xu, Danfei and Du, Yilun and Chebotar, Yevgen and Reed, Scott and Kautz, Jan and Zhu, Yuke and Fan, Linxi {``Jim''} and Jang, Joel},
  journal = {arXiv preprint arXiv:2602.15922},
  year = {2026},
  url = {https://arxiv.org/abs/2602.15922}
}

@inproceedings{kang2025physical,
  title = {How Far Is Video Generation from World Model: A Physical Law Perspective},
  author = {Kang, Bingyi and Yue, Yang and Lu, Rui and Lin, Zhijie and Zhao, Yang and Wang, Kaixin and Huang, Gao and Feng, Jiashi},
  booktitle = {Proceedings of the 42nd International Conference on Machine Learning},
  volume = {267},
  pages = {28991--29017},
  year = {2025},
  url = {https://proceedings.mlr.press/v267/kang25g.html}
}

@article{motamed2025physicsiq,
  title = {Do Generative Video Models Understand Physical Principles?},
  author = {Motamed, Saman and Culp, Laura and Swersky, Kevin and Jaini, Priyank and Geirhos, Robert},
  journal = {arXiv preprint arXiv:2501.09038},
  year = {2025},
  url = {https://arxiv.org/abs/2501.09038}
}

@inproceedings{fei2025liberoplus,
  title = {{LIBERO-Plus}: A Progressive Robustness Benchmark for Visual-Language-Action Models},
  author = {Fei, Senyu and Wang, Siyin and Shi, Junhao and Dai, Zihao and Cai, Jikun and Qian, Pengfang and Ji, Li and He, Xinzhe and Zhang, Shiduo and Fei, Zhaoye and Fu, Jinlan and Gong, Jingjing and Qiu, Xipeng},
  booktitle = {Proceedings of the IEEE/CVF Conference on Computer Vision and Pattern Recognition (CVPR)},
  pages = {38574--38583},
  year = {2026},
  url = {https://openaccess.thecvf.com/content/CVPR2026/html/Fei_LIBERO-Plus_A_Progressive_Robustness_Benchmark_for_Visual-Language-Action_Models_CVPR_2026_paper.html}
}

@article{xu2025bayesvla,
  title = {Seeing to Act, Prompting to Specify: A {Bayesian} Factorization of Vision Language Action Policy},
  author = {Xu, Kechun and Zhu, Zhenjie and Chen, Anzhe and Zhao, Shuqi and Huang, Qing and Yang, Yifei and Lu, Haojian and Xiong, Rong and Tomizuka, Masayoshi and Wang, Yue},
  journal = {arXiv preprint arXiv:2512.11218},
  year = {2025},
  url = {https://arxiv.org/abs/2512.11218}
}

@article{an2025aiflow,
  title = {{AI Flow}: Perspectives, Scenarios, and Approaches},
  author = {An, Hongjun and Hu, Wenhan and Huang, Sida and Huang, Siqi and Li, Ruanjun and Liang, Yuanzhi and Shao, Jiawei and Song, Yiliang and Wang, Zihan and Yuan, Cheng and Zhang, Chi and Zhang, Hongyuan and Zhuang, Wenhao and Li, Xuelong},
  journal = {Vicinagearth},
  volume = {3},
  eid = {1},
  year = {2026},
  doi = {10.1007/s44336-025-00031-y},
  url = {https://link.springer.com/article/10.1007/s44336-025-00031-y}
}

@article{shao2025aiflowedge,
  title = {{AI Flow} at the Network Edge},
  author = {Shao, Jiawei and Li, Xuelong},
  journal = {IEEE Network},
  volume = {40},
  number = {1},
  pages = {330--336},
  year = {2026},
  doi = {10.1109/MNET.2025.3541208},
  url = {https://ieeexplore.ieee.org/document/10884554/}
}

@article{liang2025rlvisual,
  title = {Integrating Reinforcement Learning with Visual Generative Models: Foundations and Advances},
  author = {Liang, Yuanzhi and Fang, Yijie and Li, Rui and Ni, Ziqi and Su, Ruijie and Zhang, Chi},
  journal = {Vicinagearth},
  volume = {3},
  eid = {2},
  year = {2026},
  doi = {10.1007/s44336-025-00030-z},
  url = {https://link.springer.com/article/10.1007/s44336-025-00030-z}
}

@article{zhang2024vast,
  title = {{VAST 1.0}: A Unified Framework for Controllable and Consistent Video Generation},
  author = {Zhang, Chi and Liang, Yuanzhi and Qiu, Xi and Yi, Fangqiu and Li, Xuelong},
  journal = {arXiv preprint arXiv:2412.16677},
  year = {2024},
  url = {https://arxiv.org/abs/2412.16677}
}

@article{liang2026teleboost,
  title = {{TeleBoost}: A Systematic Alignment Framework for High-Fidelity, Controllable, and Robust Video Generation},
  author = {Liang, Yuanzhi and Wu, Xuan'er and Liu, Yirui and Fang, Yijie and Fan, Yizhen and Hao, Ke and Li, Rui and Liu, Ruiying and Ni, Ziqi and Yu, Peng and Wang, Yanbo and Huang, Haibin and Weng, Qizhen and Zhang, Chi and Li, Xuelong},
  journal = {arXiv preprint arXiv:2602.07595},
  year = {2026},
  url = {https://arxiv.org/abs/2602.07595}
}

@inproceedings{ni2026vipo,
  title = {Seeing What Matters: Visual Preference Policy Optimization for Visual Generation},
  author = {Ni, Ziqi and Liang, Yuanzhi and Li, Rui and Zhou, Yi and Huang, Haibin and Zhang, Chi and Li, Xuelong},
  booktitle = {Proceedings of the IEEE/CVF Conference on Computer Vision and Pattern Recognition (CVPR)},
  pages = {27260--27269},
  month = {June},
  year = {2026},
  url = {https://openaccess.thecvf.com/content/CVPR2026/html/Ni_Seeing_What_Matters_Visual_Preference_Policy_Optimization_for_Visual_Generation_CVPR_2026_paper.html}
}

@inproceedings{liu2026bpgo,
  title = {Learning What to Trust: {Bayesian} Prior-Guided Optimization for Visual Generation},
  author = {Liu, Ruiying and Liang, Yuanzhi and Huang, Haibin and Yu, Tianshu and Zhang, Chi},
  booktitle = {Proceedings of the IEEE/CVF Conference on Computer Vision and Pattern Recognition (CVPR)},
  pages = {34408--34417},
  month = {June},
  year = {2026},
  url = {https://openaccess.thecvf.com/content/CVPR2026/html/Liu_Learning_What_to_Trust_Bayesian_Prior-Guided_Optimization_for_Visual_Generation_CVPR_2026_paper.html}
}

@article{li2026taros,
  title = {Rethinking Reward Signals in Video {GRPO}: When Scores Become Targets},
  author = {Li, Rui and Liang, Yuanzhi and Ni, Ziqi and Huang, Haibin and Zhang, Chi and Li, Xuelong},
  journal = {arXiv preprint arXiv:2511.19356},
  year = {2026},
  url = {https://arxiv.org/abs/2511.19356}
}

@article{li2026otca,
  title = {Learning to Credit the Right Steps: Objective-aware Process Optimization for Visual Generation},
  author = {Li, Rui and Hao, Ke and Liang, Yuanzhi and Huang, Haibin and Zhang, Chi and Gu, Yun and Li, Xuelong},
  journal = {arXiv preprint arXiv:2604.19234},
  year = {2026},
  url = {https://arxiv.org/abs/2604.19234}
}

@article{li2026rats,
  title = {Reward-Aware Trajectory Shaping for Few-step Visual Generation},
  author = {Li, Rui and Li, Bingyu and Liang, Yuanzhi and Huang, Haibin and Zhang, Chi and Li, Xuelong},
  journal = {arXiv preprint arXiv:2604.14910},
  year = {2026},
  url = {https://arxiv.org/abs/2604.14910}
}

@inproceedings{liang2019vrr,
  title = {{VrR-VG}: Refocusing Visually-Relevant Relationships},
  author = {Liang, Yuanzhi and Bai, Yalong and Zhang, Wei and Qian, Xueming and Zhu, Li and Mei, Tao},
  booktitle = {Proceedings of the IEEE/CVF International Conference on Computer Vision (ICCV)},
  pages = {10402--10411},
  year = {2019},
  doi = {10.1109/ICCV.2019.01050},
  url = {https://arxiv.org/abs/1902.00313}
}

@inproceedings{liang2022seeg,
  title = {{SEEG}: Semantic Energized Co-Speech Gesture Generation},
  author = {Liang, Yuanzhi and Feng, Qianyu and Zhu, Linchao and Hu, Li and Pan, Pan and Yang, Yi},
  booktitle = {Proceedings of the IEEE/CVF Conference on Computer Vision and Pattern Recognition (CVPR)},
  pages = {10463--10472},
  year = {2022},
  url = {https://openaccess.thecvf.com/content/CVPR2022/html/Liang_SEEG_Semantic_Energized_Co-Speech_Gesture_Generation_CVPR_2022_paper.html}
}

@inproceedings{liang2022elp,
  title = {A Simple Episodic Linear Probe Improves Visual Recognition in the Wild},
  author = {Liang, Yuanzhi and Zhu, Linchao and Wang, Xiaohan and Yang, Yi},
  booktitle = {Proceedings of the IEEE/CVF Conference on Computer Vision and Pattern Recognition (CVPR)},
  pages = {9549--9559},
  year = {2022},
  doi = {10.1109/CVPR52688.2022.00934},
  url = {https://openaccess.thecvf.com/content/CVPR2022/html/Liang_A_Simple_Episodic_Linear_Probe_Improves_Visual_Recognition_in_the_Wild_CVPR_2022_paper.html}
}
}

\end{document}